\documentclass[a4paper,11pt]{amsart}

\makeatletter
\def\enddoc@text{\ifx\@empty\@translators \else\par\bigskip\@settranslators\fi
  \ifx\@empty\addresses \else\null\par\vfill\@setaddresses\fi%
}
\def\@setaddresses{\par
  \nobreak \begingroup
\footnotesize
  \def\author##1{\nobreak\addvspace\bigskipamount}%
  \def\\{\unskip, \ignorespaces}%
  \interlinepenalty\@M
  \def\address##1##2{\begingroup
     \par\addvspace\bigskipamount\noindent
     \@ifnotempty{##1}{{\itshape\ignorespaces##1\unskip}\/: }%
     {\scshape\ignorespaces##2}\par\endgroup}%
  \def\curraddr##1##2{\begingroup
     \@ifnotempty{##2}{\nobreak\noindent{\itshape Current address}%
       \@ifnotempty{##1}{ (\ignorespaces##1\unskip)}\/:\space
       ##2\par\endgroup}}%
  \def\email##1##2{\begingroup
     \@ifnotempty{##2}{\nobreak\noindent{\itshape E-mail address}%
       \@ifnotempty{##1}{ (\ignorespaces##1\unskip)}\/:\space
       \ttfamily##2\par\endgroup}}%
  \def\urladdr##1##2{\begingroup
     \@ifnotempty{##2}{\nobreak\indent{\itshape URL}%
       \@ifnotempty{##1}{ (\ignorespaces##1\unskip)}\/:\space
       \ttfamily##2\par\endgroup}}%
  \addresses
  \endgroup
}
\def\specialsection{\@startsection{section}{1}%
   \z@{\linespacing\@plus\linespacing}{.5\linespacing}%
   {\normalsize\itshape\centering}}
\def\section{\@startsection{section}{1}%
   \z@{\linespacing\@plus\linespacing}{.5\linespacing}%
   {\normalfont\scshape\centering}}
\def\subsection{\@startsection{subsection}{2}%
   \z@{.7\linespacing\@plus.5\linespacing}{-.7em}%
   {\normalfont\bfseries}}
\def\subsubsection{\@startsection{subsubsection}{3}%
   \z@{.5\linespacing\@plus.5\linespacing}{-.7em}%
   {\normalfont\itshape}}
\def\paragraph{\@startsection{paragraph}{4}%
   \z@{.5\linespacing\@plus.3\linespacing}{-.5em}%
   {\normalfont\itshape}}
\def\subparagraph{\@startsection{subparagraph}{5}%
   \z@{.5\linespacing\@plus.3\linespacing}{-.5em}%
   {\normalfont\itshape}}
\def\th@plain{%
  \thm@headfont{\scshape}
  \let\thmhead\thmhead@plain \let\swappedhead\swappedhead@plain
  \thm@preskip.5\baselineskip\@plus.2\baselineskip
                                    \@minus.2\baselineskip
  \thm@postskip\thm@preskip
  \itshape
}
\def\th@definition{%
  \thm@headfont{\bfseries}
  \let\thmhead\thmhead@plain \let\swappedhead\swappedhead@plain
  \thm@preskip.5\baselineskip\@plus.2\baselineskip
                                    \@minus.2\baselineskip
  \thm@postskip\thm@preskip
  \upshape
}
\def\th@remark{%
  \thm@headfont{\itshape}
  \let\thmhead\thmhead@plain \let\swappedhead\swappedhead@plain
  \thm@preskip.5\baselineskip\@plus.2\baselineskip
                                    \@minus.2\baselineskip
  \thm@postskip\thm@preskip
  \upshape
}
\renewenvironment{proof}[1][\proofname]{\par
  \normalfont
  \topsep6\p@\@plus6\p@ \trivlist
  \item[\hskip\labelsep\itshape
    #1\@addpunct{.}]\ignorespaces
}{%
  \qed\endtrivlist
}
\makeatother

\usepackage{natbib}
\usepackage{amssymb,amsmath,amsthm}
\usepackage{latexsym}

\bibpunct[$\!$]{(}{)}{,}{a}{}{,}

\newcommand{\A}{\ensuremath{{\Bbb A}}}
\newcommand{\B}{\ensuremath{{\Bbb B}}}
\newcommand{\Lang}{\ensuremath{\mathcal{L}}}

\newcommand{\F}{\ensuremath{\mathcal{F}}}

\newcommand{\kon}{\raisebox{.5ex}{\ensuremath{\smallfrown}}}
\newcommand{\traRa}{\ensuremath{\stackrel{\ast}{\Rightarrow}\,}}
\newcommand{\tuple}[2][n]{\ensuremath{#2_1,\dots,#2_{#1}}}

\let\func=\longrightarrow
\let\ra=\rightarrow
\let\Ra=\Rightarrow

\newtheorem{theorem}{Theorem}
\newtheorem{claim}[theorem]{Claim}

\theoremstyle{definition}
\newtheorem{definition}{Definition}
\theoremstyle{remark}
\newtheorem{example}{Example}
\newtheorem{remark}{Remark}

\begin{document}

\title[Adjunction as Substitution]{Adjunction as Substitution\\[5pt]
 An Algebraic Formulation of Regular,
 Context-Free and Tree Adjoining Languages}
\author{Uwe M\"onnich}
\thanks{To appear in: Proceedings \emph{Formal Grammar Conference},
 Aix-en-Provence, Aug. 10, 1997}
\address{T\"ubingen University\\
SfS\\
Wilhelmstr.~113\\
D-72074 T\"ubingen}
\email{uwe.moennich@uni-tuebingen.de}
\maketitle

\nocite{ES77:IO-OI1}

\section{Introduction}
There have been many attempts to give a coherent formulation of a
hierarchical progression that would lead to a refined partition of the
vast area stretching from the context-free to the context-sensitive
languages. The purpose of this note is to describe a theory that seems
to afford a promising method of interpreting the tree adjoining
languages as the natural third step in a hierarchy that starts with
the regular and the context-free languages.

The formulation of this theory is inspired by two sources: the
categorical concept of an algebraic theory and the powerful tool of macro
variables which is well established within the framework of program
schemes. The rough idea is that according to their intended
interpretation objects of algebraic theories are sets of derived
operations and that macro variables range over these sets. Guided by this
conception we show how monadic macro variables provide a perspective from
which both the context-free and the tree adjoining languages become
realizations of the same general notion when the relevant underlying
algebra is specified. The context-free languages are determined by an
inductive process in which monadic macro variables are replaced by
derived operations of an algebra all of whose operations are unary. In
the case of the tree adjoining languages the same substitution process is
applied to derived operations over an arbitrary algebra.

The central notion in this account is that of a higher-order
substitution. Whereas in traditional presentations of rule systems for
language families the emphasis has been on a first-order
substitution process in which auxiliary variables are replaced by
elements of the carrier of the proper algebra---concatenations of
terminal and auxiliary category symbols in the string case---we lift this
process to the level of operations defined on the elements of the carrier
of the algebra. Our own view is that this change of emphasis provides the
adequate platform for a better understanding of the operation of
adjunction. To put it in a nutshell: Adjoining is not a first-order, but
a second-order substitution operation.

This is not the first place that macro productions are put to use outside
the field of program schemes. It has been known since the pioneering work
of M.J.\ Fischer that macro grammars (without the restriction to monadic
variables) are weakly equivalent to indexed  grammars. The paper by
Maibaum also contains one direction of the equivalence between context-free
grammars and monadic extensions of regular grammars. The other
direction seems to be part of the folklore wisdom pertaining to tree
language theory.

The greatest advantage that comes with a macro-like presentation derives
from an explicit notation for derived operators in an algebra. As was
emphasized above, macro variables range over complex or
derived operations. These derived operations are formed by functional
composition, starting from the primitive operations. There is an
alternative conception of this compositionally closed family of
functions  as an algebra whose carrier is the disjoint union of the
derived operations and whose only primitive operations are just the
suitably typed composition functions. This conception views operations as
elements of a many-sorted algebra. The corresponding shift on the
symbolic level "nominalizes" the second-order macro variables and maps
them into object variables. As a result of this translational shift a
system with macro-like productions becomes an instance of a regular tree
grammar. This latter grammar format provides the class of structures that
fall under the purview of logical techniques that have been successfully
applied to problems surrounding the metatheory of several linguistic
models. The algebraic approach sketched above will enable us to apply
these techniques to structural properties of grammatical formalisms that
properly include the range of mildly context-sensitive devices.

In the following we will first introduce the basic notion of a
context-free tree grammar (Section 2). Then, in Section 3, the
equivalence between context-free string languages and context-free
tree languages over monadic alphabets is established. Section 4 is
devoted to the presentation of the tree adjoining languages as
languages generated by context-free tree grammars where all the
functional symbols are monadic. The concluding section points to
the model theoretic techniques that are applicable to the kinds of
tree producing systems underlying the constructions in this paper.

\section{Preliminaries}

The purpose of this section is to fix notations and to present
definitions related to tree grammars.

\begin{definition} A \textbf{ranked alphabet} or \textbf{signature}
$\Sigma$ is an indexed family $\Sigma_n(n \in {\Bbb N})$ of disjoint
sets. A symbol in $\Sigma_n$ is called an operator of \textbf{rank}
$n$. If $n=0$ then $f \in \Sigma_0$ is also called a
\textbf{constant}.
\end{definition}

\begin{example}
  \hspace*{2cm}
\begin{enumerate}
  \renewcommand{\labelenumi}{\alph{enumi})}
  \renewcommand{\theenumi}{\alph{enumi}}
\item $\Sigma_0 = \{ \varepsilon \}\cup V$ \hspace*{.5cm} $\Sigma_2 =
  \{\kon\}$\\
  Single--sorted signature of semi--groups, extended by a finite set of
  constants $V$.
\item $\Sigma_0 = \{ \varepsilon \}$ \hspace*{.5cm} $\Sigma_1 = \{a \,
  | \, a\in V  \}$\\
  Single--sorted signature of a \emph{monadic} algebra.
\end{enumerate}
\end{example}

\begin{definition} For a ranked alphabet $\Sigma$, we denote by
$T(\Sigma)$ the set of \textbf{trees} over $\Sigma$. $T(\Sigma)$
is inductively defined as follows:
\begin{enumerate}
  \renewcommand{\labelenumi}{(\roman{enumi})}
\item $\Sigma_0 \subseteq T(\Sigma)$
\item If $f \in \Sigma_n$ and $t_i \in T(\Sigma)$ for $i=1,\ldots,n$\\
~\hfill$f(t_1,\dots,t_n) \in T(\Sigma)$\hfill~
\end{enumerate}
\end{definition}

We fix an indexed set $X=\{x_1, x_2,\ldots\}$ of \emph{variables}
and denote by $X_n$ the subset $\{x_1,\dots, x_n\}$. Variables are
considered to be constants, i.e. operators of rank $0$.
 For a
ranked alphabet $\Sigma$ the family $T(\Sigma,X)$ is defined to be
$T(\Sigma(X))$, where $\Sigma(X)$ is the ranked alphabet with
$\Sigma(X)_0=\Sigma_0\cup X$ and $\Sigma(X)_n=\Sigma_n$ for every
$n\neq 0$. For $Y\subseteq X$ a tree $t\in T(\Sigma,X)$ is called
\emph{linear} in $Y$ if no variable in $Y$ has more than one
occurrence in $t$. A subset $L$ of $T(\Sigma)$ is called a
\emph{tree language} over $\Sigma$.

Having described the tree terms, it remains to specify the central
notion of an algebra and to give a precise definition of the way in
which the operator symbols induce an operation on an algebra.

\begin{definition} Suppose that $\Sigma$ is a ranked alphabet. A
$\Sigma$-\textbf{algebra} ${\Bbb A}$ is a pair $ {\Bbb A}=(A,(f_{\Bbb A})_{f\in\Sigma})$ where
the set $A$ is the \textbf{domain} or \textbf{carrier} of the
algebra and for each operator $f\in\Sigma_n, f_{\Bbb A}: A^n\rightarrow A$
is an operation of arity $n$ on $A$.
\end{definition}
Different algebras, defined over the same operator domain, are related
to each other if there exists a mapping between their carriers that is
compatible with the basic structural operations.

\begin{definition}
  A $\Sigma$-\textbf{homomorphism} of $\Sigma$-algebras $h: \A \func \B$
  is a function $h: A \func B$,
  such that for every operator $f$ of rank $n$
 \[  h(f_\A(\tuple{a})) =
  f_\B(h(a_1),\dots,h(a_n))\] for every $n$-tuple
  $(\tuple{a})\in A^n$.
\end{definition}

The set of trees $T(\Sigma,X)$ can be made into a $\Sigma$-algebra
by defining the operations in the following way. For every $f$ in
$\Sigma_n$, for every $(t_1,\ldots,t_n)$ in $T(\Sigma,X)^n$
\[ f_{{\Bbb T}(\Sigma,X)}(\tuple{t}) = f(\tuple{t}).\]

Every variable-free tree $t\in T(\Sigma)$ has a  value in every
$\Sigma$-algebra ${\Bbb A}$. It is the value at $t$ of the unique
homomorphism $h: {\Bbb T}(\Sigma)\rightarrow {\Bbb A}$.

The existence of a unique homomorphism from the $\Sigma$-algebra of
trees into an arbitrary $\Sigma$-algebra ${\Bbb A}$ provides also
 the basis
for the view that regards the elements of $T(\Sigma,X_n)$ as
\emph{derived operations}. Each tree $t \in T(\Sigma, X_n)$
induces an $n$-ary function

\[ t_{\Bbb A}: A^n\rightarrow A \]

The meaning of this function $t_{\Bbb A}$ is defined in the following way.
For every $(a_1,\ldots,a_n)\in A^n$
$t_{\Bbb A}(a_1,\ldots,a_n)=\hat{a}(t)$,
where $\hat{a}: {\Bbb T}(\Sigma,X_n)\rightarrow {\Bbb A}$ is the unique
homomorphism with $\hat{a}(x_i)=a_i$.

In the particular case where ${\Bbb A}$ is the $\Sigma$-algebra
${\Bbb T}(\Sigma,X_m)$ of trees over $\Sigma$ that contain at most
variables from $X_m=\{x_1,\ldots,x_m\}$ at their leaves the unique
 homomorphism extending
the assignment of a tree $t_i\in T(\Sigma, X_m)$ to the variable
$x_i$ in $X_n$ acts as a substitution:
\[ t_{{\Bbb T}(\Sigma,X_m)}(\tuple{t}) = t[\tuple{t}]\]
where the last tree indicates the result of substituting $t_i$ for
$x_i$ in $t$.

Let $V$ be an ordinary finite alphabet. It gives rise to a
\emph{monadic} signature $\Sigma_V$ if all the members of $V$ are
assigned rank one and a new symbol $\varepsilon$ is added as the single
constant of rank zero. As was pointed out in the previous
paragraph, there is a unique homomorphism from the trees over
$\Sigma_V$, considered as the carrier of a $\Sigma_V$-algebra, to any
other algebra of the same type. In particular, there is a
homomorphism into $V^*$ when every $a \in V$, i.e.\ $a \in
(\Sigma_{V})_1$, is interpreted as left-concatenation with the symbol
$a$ and when $\varepsilon$ is interpreted as the constant string of length
zero. This homomorphism is actually an isomorphism between
$T(\Sigma_V)$ and $V^*$ \cite[cf.][]{maibaum74}. This isomorphism will play
an important role in the next section on context-free string
languages.

We now turn to introducing the notion of a context-free tree grammar (cftg).
This type of grammar is related to the type of grammars that were defined
by \cite{fischer68} and were called macro grammars by him. Context-free
tree grammars constitute an algebraic generalization of macro grammars
since the use of macro-like productions served the purpose of making
simultaneous string copying a primitive operation.

\begin{definition}
  A \textbf{context--free tree grammar} $G = \langle \Sigma, \F, S, P
  \rangle$ is a 4-tuple, where $\Sigma$ is a finite ranked alphabet of
  \emph{terminals}, $\F$ is a finite ranked alphabet of
  \emph{nonterminals} disjoint from $\Sigma$, $S\in \F$ is the start
  symbol of rank $0$ and $P$ is a finite set of rules of the form
\[
F(x_1,\dots,x_m) \ra t \hspace*{1cm} (n \geq 0)
\]
where $F\in \F_m$ and $t\in T(\Sigma\cup\F,X_m)$. Recall that $X$ is
here assumed to be a set of variables $X= \{x_1,x_2,\dots\}$ and $X_m =
\{x_1,\dots,x_m\}$.
\end{definition}

In the conventional case of context-free string grammars the class
of languages generated by unrestricted derivations is identical to
the class of languages generated by left-most derivations (or by
right-most derivations, for that matter). Such an equivalence
cannot be proved in the tree case. While this is true for
cftg's in general, the difference between the
derivation modes has no effect on the special category of monadic
linear tree grammars, which will be our only concern in the sequel.

The following definition records the unrestricted derivation mode.
\begin{definition}
  Let $G = \langle \Sigma, \F, S, P \rangle$ be a cftg and let $t,t' \in T(\Sigma\cup\F)$.  $t'$ is
  \textbf{directly derivable}  from $t$ ($
  t \Ra t' $) if there is a tree $t_0 \in T(\Sigma\cup\F,X_{1})$
  containing exactly \emph{one} occurrence of $x_{1}$, a corresponding
  rule $F(x_1,\dots,x_m) \ra t''$, and trees $t_1,\dots,t_m \in
  T(\Sigma)$ such that
\begin{eqnarray*}
  t & = & t_0[F(t_1,\dots,t_m)]\\
  t' & = & t_0[t''[t_1,\dots,t_m]]
\end{eqnarray*}
$t'$ is obtained from $t$ by replacing an occurrence of a subtree
$F(t_1,\dots,t_m)$ by the tree $t''[t_1,\dots,t_m]$.

Recall from above that for $m,n\geq 0$, $t\in
T(\Sigma,X_m)$ and $t_1,\dots,t_m \in T(\Sigma,X_n)$ $ t[t_1,\dots,t_m]
$ denotes the result of \textbf{substituting} $t_i$ for $x_i$ in $t$.
Observe that $t[t_1,\dots,t_m]$ is in $T(\Sigma,X_n)$.

As is customary \traRa\ denotes the transitive-reflexive closure of
$\Ra$.
\end{definition}

\begin{definition}
  Suppose $G = \langle \Sigma, \F, S, P \rangle$ is a cftg. We call
\[
\mathcal{L}(G,S) = \{ t\in T(\Sigma) \, | \, S \traRa t \}
\]
the \textbf{context-free tree language} generated by $G$ from
$S$.
\end{definition}

We reserve a special definition for the case where $\F$ contains only
function symbols of rank zero.

\begin{definition}
  A \textbf{regular tree grammar} is a tuple $G = \langle \Sigma, \F, S,
  P \rangle$, where $\Sigma$ is a finite ranked alphabet of terminals,
  $\F$ is a finite alphabet of function or nonterminal symbols of rank
  zero, $S\in\F$ is the start symbol and $P\subseteq \F\times
  T(\Sigma\cup\F)$ is a finite set of productions.  The \textbf{regular
    tree language} generated by $G$ is
\[ \Lang = \{ t\in T(\Sigma)\, | \, S \traRa t \}
\]
\end{definition}

Note that in the case of regular grammars the analogy with the
conventional string theory goes through. There is an equivalence of the
unrestricted, the rightmost and the leftmost derivation modes where the
terms 'rightmost' and 'leftmost' are to be understood with respect to
the linear order of the leaves forming the frontier of a tree in a
derivation step.

\section{Monadic Trees}
Let us view grammars as a
mechanism in which local transformations on trees can be performed in a
precise way. The central ingredient of a grammar is a finite set of
productions, where each production is a pair of trees. Such a set of
productions determines a binary relation on trees such that two trees
$t$ and $t'$ stand in that relation if $t'$ is the result of removing in
$t$ an occurrence of a first component in a production pair and
replacing it by the second component of the same pair. The simplest type
of such a replacement is defined by a production that specifies the
substitution of a single-node tree $t_0$ by another tree $t_{1}$. Two
trees $t$ and $t'$ satisfy the relation determined by this simple
production if the tree $t'$ differs from the tree $t$ in having a
subtree $t_{1}$ that is rooted at an occurrence of a leaf node $t_{0}$
in $t$. In slightly different terminology, productions of this kind
incorporate instructions to rewrite auxiliary variables as a complex
symbol that, autonomously, stands for an element of a tree algebra. As
long as the carrier of a tree algebra is made of constant tree terms the
process whereby nullary variables are replaced by trees is analogous to
what happens in string languages when a nonterminal auxiliary symbol is
rewritten as a string of terminal and non-terminal symbols,
independently of the context in which it occurs. The situation changes
dramatically if the carrier of the algebra is made of symbolic
counterparts of derived operations and the variables in production rules
range over such second-level entities.  The following example
 illustrates the gain
in generative power to be expected from production systems determining
relations among trees that derive from second-order substitutions of
operators rather than constants.

\begin{example} Let us assume that we are dealing with a vocabulary
$V$ that contains the symbols $a$ and $b$ as its only members.
Trees over the associated monadic signature $\Sigma=\Sigma_0\cup
\Sigma_1$, where $\Sigma_0=\{\varepsilon\}$ and $\Sigma_1=\{a,b\}$
are arbitrary sequences of applications of the operators $a$ and
$b$ to the constant $\varepsilon$. As will be recalled from the
preceding section, trees in $T(\Sigma,X)$ can be regarded as
derived operators. Due to the fact that $\Sigma$ is a
\emph{monadic} signature these trees
 may not contain
more than a single variable. Auxiliary symbols ranging over this
domain take therefore as values monadic or constant derived
operators. When monadic auxiliary symbols
 appear in productions this means that they behave the
way that nullary auxiliary symbols do except for the fact that their
argument has to be inserted into the unique variable slot of their
replacing derived operator. After these explanations it should be
obvious that the following grammar over the \emph{monadic}
signature $\Sigma$ generates the context-free string language $\{
a^nb^n\}$ when combined with the unique isomorphism mentioned above:

\begin{gather*}
G = \langle \Sigma, \F, S, P \rangle
\\
\begin{array}{lclclcl}
  \F_0 & = & \{S\} & \hspace*{1cm} & \F_1 & = & \{ F\}
\end{array}
\\
P = \left\{ \begin{array}{lcl}
    S & \!\ra\! & F(\varepsilon)\, |\,\varepsilon  ,\\
    F(x) &\! \ra\! & a(F(b(x))) \, | \, a(b(x))
\end{array}\right\}
\\
\mathcal{L}(G,S) = \{
\overbrace{a(a\dots}^n\overbrace{(b(b\dots}^n(\varepsilon)\dots)\}
\end{gather*}

\end{example}

The grammar in the last example illustrates for the language
$L=\{ a^nb^n\}$ a transformation that can be applied to the grammar of any
given context-free language.

\begin{claim}
For every finite alphabet $V$, the
context-free tree subsets of $T(\Sigma_V)$ (where $\Sigma_{V}$ is the monadic
signature induced by $V$) have exactly the context-free string languages
as values of the unique isomorphism that maps trees in $T(\Sigma_V)$
into strings in $V^\ast$ and interprets the elements of $V$ as unary
operations that concatenate for the left with their argument.
\end{claim}

\begin{proof}
 Let $G=(N,V,S,P)$ be an arbitrary context-free string
grammar, where $N$ is a finite set of nonterminal symbols, $V$ the
finite terminal alphabet disjoint from $N, S \in N$ the initial or
start symbol and $P\subseteq N\times(N\cup V)^*$ a finite set of
productions. We associate with $G$ an equivalent context-free tree
grammar $G'=(\Sigma_V, \mathcal{F}, S, P')$ in the following way.
We let $\Sigma_V$ be the monadic ranked alphabet corresponding to
$V$. We let $\mathcal{F}=\mathcal{F}_0\cup\mathcal{F}_1$ where
$\mathcal{F}_0=\{S'\}$ and $\mathcal{F}_1=N$, be the ranked alphabet
of nonterminals. The new set of productions $P'$ is obtained as
follows. If $A\rightarrow w_1 \dots w_n$ is in $P$, where $w_i \in
N\cup V$ then $A(x)\rightarrow w_1(w_2(\dots(w_n(x)\dots)$ is in
$P'$. In addition, we have a rule for the new start symbol:
$S'\rightarrow S(\varepsilon)$. It should be clear from the
preceding example that $L(G)=h(L(G'))$ where $h$ denotes the unique
isomorphism between $V^*$ and $T(\Sigma_V)$. For the other direction
assume that $G=(\Sigma,\mathcal{F},S,P)$ is an arbitrary
cftg over the \emph{monadic} terminal
alphabet $\Sigma$. We associate with $G$ an equivalent context-free
string grammar $G'=(\mathcal{F},\Sigma,S,P')$ in the following way.
$\mathcal{F}$ and $\Sigma$ are regarded as unranked sets and play
the role of nonterminal and terminal alphabets, respectively. The
set of productions $P'$ is obatined from the paths in the members
of $P$. More precisely. Let $F(x_1,\dots,x_n)\rightarrow t$ be a
member of $P$ with $t\in T(\Sigma\cup\mathcal{F},X_n)$. Then each
$F\rightarrow t'$ is in $P'$, wher $t'$ is the concatenation of
 symbols from $\Sigma$ and
$\mathcal{F}$ that label a path through $t$. It should be obvious
that nonterminals in $\mathcal{F}_n$ for $n>1$ have to be
ultimately replaced by linear or constant monadic trees in a
derivation according to the original cftg $G$
if they are to contribute to the generated language at all.
Spurious nonterminals are retained in the process of substituting
the path sets for the trees on the right hand sides of the rules in
$P$, but they still do not affect the string languages specified by
$G'$. Indeed, under the unique isomorphism between $V^*$ and
$T(\Sigma)$ it holds that $G$ and $G'$ generate the same language.
\end{proof}
Let $V$ be an ordinary finite alphabet and $\Sigma_V$ ist
associated monadic signature. Recall that all nonterminals in a
regular tree grammar are of arity zero. An arbitrary rule in a
regular tree grammar over the monadic signature $\Sigma_V$ can
therefore assume only one of the following two forms:
\begin{align*}
A\rightarrow& a_1(a_2(\dots a_n(\varepsilon)\dots)\\
A\rightarrow& a_1(a_2(\dots a_n(B)\dots)
\end{align*}
where $a_i \in (\Sigma_V)_1$, i.e. the $a_i$'s are the monadic
terminal operators corresponding to the members of $V$ and $A$ and
$B$ are two nonterminals of rank zero. Under the unique isomorphism
that relates strings with monadic "vertical" trees the two rule
formats above correspond exactly to the two types of rules in
regular string grammars. This implies that a language $L$ generated
by a regular tree grammar over a unary ranked alphabet is,
interpreted as a string language, the language generated by the
corresponding regular string grammar. Based on the claim about the
relationship between context-free string grammars and context-free
tree grammars over a finite vocabulary and its associated monadic
signature, respectively, we have therefore established the result,
announced in the introduction, that the increase in generative power
which characterizes the shift from the regular to the context-free
languages, is due to the transgression from a nullary, first-order
substitution process to its higher-order analogue.

\section{Adjoining Trees}

Very early in the development of (regular) tree grammars it was realized
that there exists a close relationship between the families of trees
generated by tree grammars and the family of context-free string
languages. This fundamental fact is best described by looking at it from
the perspective on trees that views them as symbolic representations of
values in arbitrary domains.

\begin{definition}
  Suppose $\Sigma$ is a ranked alphabet. We call
  \textbf{yield} or \textbf{frontier} the unique homomorphism $y$ that
  interprets every operator in  $\Sigma_{n}$
  as the $n$-ary operation of concatenation. More precisely
\[
\begin{array}{rcll}
  y(f) & = & f & \text{ for } f \in  \Sigma_0\\
  y(f(\tuple{t})) & = & y(t_1)\dots y(t_n) & \text{ for } f
  \in \Sigma_n \text{ and } t_i \in T(\Sigma)
\end{array}
\]
\end{definition}

\smallskip\noindent\textbf{Fact} \textit{
  A (string) language is context-free iff it is the yield of a regular
  tree language.}
\smallskip

As was shown in the last section, the addition of macro variables
or n-ary nonterminals $(n>0)$ increases the generative power of
cftg's over monadic alphabets considerably. In
addition, the transformation of an arbitrary context-free tree
grammar over a monadic alphabet into an equivalent context-free
string grammar has shown that it is only the monadic nonterminals
that are operative in the derivation process with a terminal
outcome, i.e. a (tree-)expression over the terminal alphabet. The
obvious question that presents itself in this situation is whether the
addition of monadic nonterminals to a regular grammar over an
arbitrary signature leads to a langugae family that has already
been introduced for independent reasons.

\begin{example}
 Let $G=(\Sigma,\mathcal{F},S',P)$ be a context-free tree
grammar with the following specifications:
\begin{gather*}
\begin{split}
\Sigma=\Sigma_0\cup\Sigma_3 &\text{ where } \Sigma_0=\{a,b,c,d\}
\text{ and } \Sigma_3=\{S\}\\
\mathcal{F}=\mathcal{F}_0\cup\mathcal{F}_1 &\text{ where }
 \mathcal{F}_0=\{S'\} \text{ and } \mathcal{F}_1=\{\bar{S}\}
\end{split}\\
P=\{S'\rightarrow \bar{S}(\varepsilon),\quad \bar{S}(x) \rightarrow x,\quad
\bar{S}(x)\rightarrow S(a,\bar{S}(S(b,x,c)),d)\}
\end{gather*}

In tree form the last rule has the following shape:

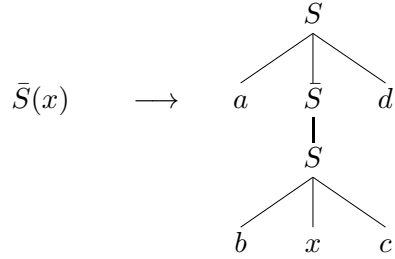
\begin{center}
\parbox{.5\linewidth}{%
{\parskip=0pt\offinterlineskip%
\hskip 10.50em\hbox to 0.50em{\hss{$S$}\hss}%
\vrule width0em height1.972ex depth0.812ex\par\penalty10000
\hskip 10.75em\raise1.500ex\hbox{\special{ps:currentpoint /momabby exch def /momabbx exch def}}%
\vrule width0em height1.500ex depth0.500ex\par\penalty10000
\hskip 8.25em\lower0.500ex\hbox{
}%
\hskip 2.50em\lower0.500ex\hbox{
}%
\hskip 2.50em\lower0.500ex\hbox{
}%
\vrule width0em height1.500ex depth0.500ex\par\penalty10000
\hskip 0.00em\hbox to 3.00em{\hss{~$\bar{S}(x)$}\hss}%
\hskip 2.00em\hbox to 1.00em{\hss{$\longrightarrow$~~}\hss}%
\hskip 2.00em\hbox to 0.50em{\hss{$a$}\hss}%
\hskip 2.00em\hbox to 0.50em{\hss{$\bar{S}$}\hss}%
\hskip 2.00em\hbox to 0.50em{\hss{$d$}\hss}%
\vrule width0em height1.972ex depth0.812ex\par\penalty10000
\hskip 10.75em\vrule width.04em%
\vrule width0em height1.500ex depth0.500ex\par\penalty10000
\hskip 10.50em\hbox to 0.50em{\hss{$S$}\hss}%
\vrule width0em height1.972ex depth0.812ex\par\penalty10000
\hskip 10.75em\raise1.500ex\hbox{\special{ps:currentpoint /momabfy exch def /momabfx exch def}}%
\vrule width0em height1.500ex depth0.500ex\par\penalty10000
\hskip 8.25em\lower0.500ex\hbox{
}%
\hskip 2.50em\lower0.500ex\hbox{
}%
\hskip 2.50em\lower0.500ex\hbox{
}%
\vrule width0em height1.500ex depth0.500ex\par\penalty10000
\hskip 8.00em\hbox to 0.50em{\hss{$b$}\hss}%
\hskip 2.00em\hbox to 0.50em{\hss{$x$}\hss}%
\hskip 2.00em\hbox to 0.50em{\hss{$c$}\hss}%
\vrule width0em height1.972ex depth0.812ex\par}%
}
\end{center}

This grammar generates the string language $\{a^nb^nc^nd^n\}$, i.e.,
$y(L(G))=\{a^nb^nc^nd^n\}$.
\end{example}

Apart from minor notational modifications the grammar in the last
example corresponds to a tree adjoining grammar which serves as an
illustration in the paper by \cite{vijay-shankerWeir92}. Note that
apart from the start symbol the only other nonterminal is of arity
one. As was the case in connection with the context-free string
languages, the preceding example is a particular instance of the
general situation. The tree adjoining languages correspond  to the
context-free tree languages that are specified by rule systems with
nonterminals of arity at most one.

We shall prove the equivalence between tree adjoining languages and
context-free tree languages with a monadic rule system for the case
of tree adjoining grammars with a restricted set of adjunction constraints.

The device of \emph{adjunction constraints} was introduced by A.~Joshi
for the purpose of specifying which auxiliary trees can be adjoined
at a given node. The distinction in cftg's between
 functional nonterminal and terminal labels corresponds to the
distinction between nonterminals with an obligatory and a null
adjunction constraint in an initial or an auxiliary tree of a
tree adjoining grammar. In the theorem below we shall cover this
particular distinction. The proof can be easily adapted to the general
case.

\begin{definition} A \textbf{tree adjoining grammar} (tag) $G$ is a
quintuple $G=(V,N,S,{\Bbb I},{\Bbb A})$, where (i) $V$ is a finite set of
terminal symbols, (ii) $N$ is a finite set of nonterminal symbols
disjoint from $V$, (iii) $S\in N$ is the distinguished start symbol,
(iv) ${\Bbb I}$ is a finite set of trees, called \textbf{initial trees},
whose root label is $S$ and (v) ${\Bbb A}$ is a finite set of
trees, called \textbf{auxiliary trees}, that have a distinguished
frontier node, called \textbf{foot node}, whose label is identical
to the label of the root node. Only frontier nodes in ${\Bbb I}$ or
${\Bbb A}$ are labelled by terminal symbols. Nonterminal
symbols can label both frontier and interior nodes. The case where a
node is specified by an obligatory adjunction constraint is indicated
by a bar over the node label.
\end{definition}

\begin{definition} Let $G=(V,N,S,{\Bbb I},{\Bbb A})$ be a tag and let $t,t'$ be
two trees with node labels from $V$ and $N$. $t'$ is
\textbf{directly derivable} from $t\; (t\Rightarrow t')$ if there is
a tree $t_0$ containing exactly \emph{one} occurrence of a new
leaf label $\zeta$, an auxiliary tree $t_A$ with root and foot node
label $A$ such that
\begin{align*}
t=&t_0(t''/\zeta)\\
t'=&t_0(t_A(t''/A)/\zeta)
\end{align*}
where the root of $t''$ is labelled by $\bar{A}$ and where the notation
$t_1(t_2/X)$ indicates the result of substituting the tree $t_2$
for a particular leaf node of $t_1$ which is labelled by $X$.
In this derivation step the adjunction constraint of the foot node in
$t_A$ replaces the constraint on the root of $t''$.
\end{definition}

\begin{definition} Suppose $G=(V,N,S,{\Bbb I},{\Bbb A})$ is a tag. The
\textbf{tree language} $T(G)$ generated by $G$ is defined as
\[T(G)=\{t|t_i\traRa t \text{ for some }t_i \in {\Bbb I} \text{ and }
t \text{ has no obligatory node labels}\}\]
The \emph{string language}, $L(G)$, generated by $G$ is defined
as
\[L(G)=\{\emph{frontier}(t)|t \in T(G)\text{ and }\emph{frontier}(t) \in
V^*\}\]
\end{definition}

The following theorem shows how to construct a weakly equivalent
tag $G'$ for a given cftg $G$ with contains
only nullary and monadic nonterminals and vice versa.

\begin{theorem} The classes of string languages generated by tree
adjoining grammars and by cftg's with
nonterminals of arity at most one and linear production rules are
the same.
\end{theorem}
\begin{remark}
 The linearity constraint mirrors the requirement of a
distinguished foot node in the auxiliary trees.
\end{remark}
\begin{proof}
Let $G=(V,N,S,{\Bbb I},{\Bbb A})$ be a tag. We construct an
equivalent cftg
$G'=(\Sigma,\mathcal{F},S',P)$, where we let $\Sigma_0=V$ and
$\Sigma_n=\{A_n|A \in N$ and $A$ labels a node in ${\Bbb I}$ or
${\Bbb A}$ with $n$ daughters$\}$, $\mathcal{F}_0=\{A'|A\in N\}$ and
$\mathcal{F}_1=\{\bar{A}|A\in N\}$ and we let $P$ consist of the
following productions:
\begin{gather*}
\begin{split}
S'\rightarrow\;& t_i^* \text{ for }t_i\in {\Bbb I}\\
\bar{A}(x)\rightarrow\; & t^*_a(x/A) \text{ for }t_a \in {\Bbb A}
\text{ with }A\in N \text{ labelling root and foot node, and }\\
&\phantom{ t^*_a(x/A)}\text{ carrying the null adjunction constraint
  at the foot node}\\
\bar{A}(x)\rightarrow\;& t^*_a(\bar{A}(x)/\bar{A}) \text{ otherwise}
\end{split}
\end{gather*}
The trees $t^*$ are the result of replacing every non-foot node with
a nonterminal label $A$ carrying the obligatory adjunction constraint
 by the branch consisting of the node
labelled $\bar{A}$ immediately dominating the node labelled $A$ and of
replacing every nonterminal non-foot leaf label $A$. In the
construction of the rule set $P$ we have suppressed the numerical
subscripts of the terminal symbols originating from the alphabet
$N$. A straightforward, but tedious inductive argument will show
that $G'$ is weakly equivalent with $G$ and even generates the same
tree language.

 Suppose now that $G=(\Sigma,\mathcal{F},S,P)$ is a
cftg satisfying the constraints in the
statement of the theorem. An equivalent tag
$G'=(V,N,S',{\Bbb I},{\Bbb A})$ is obtained in the following
manner. Let $V=\Sigma_0$ and let $N=\bigcup_{n>0} \Sigma_n \cup
\mathcal{F}\times P$.
We assume that $S'$ is a new start symbol that does not occur in
$\Sigma$ or $\mathcal{F}$. $\Bbb I$ consists of the result of
performing the following operations on the right-hand sides of the
nonterminal $S$ in $P$:  Occurrences of nonterminal labels $F$ are
replaced by bared pairs $\overline{(F,p)}$ such that $p$ is an element in $P$
rewriting $F$. The results of these substitutions are extended by a
new root labelled $S'$. The set of auxiliary trees $\Bbb{A}$
is obtained by the same substitution operation performed this time
on the right-hand sides of nonterminals other than $S$. The
elements in $\Bbb{A}$, too, are extended by a new root which
consists of the unbarred pair $(F,p)$ of the replaced nonterminal
$F$ and the relevant rule label $p$ that licenses the particular
replacement of $F$ by the daughter tree of the new root label. That
very same root label is also substituted for the variable $x$.
These new root and foot labels do no real work. They are introduced
for the sole purpose of complying with the specifications of a tree
adjoining grammar. The new grammar $G'$ is again, apart from minor
notational modifications, identical to its "source" and it should
be obvious that both generate the same string language.

\end{proof}

\section{Conclusion}

As mentioned in the introduction, one of our motivations for
looking into the relationship between tree adjoining and
context-free tree grammars has been the intimate connection between
descriptive complexity theory and algebraic language theory. It has
been known for nearly 30 years that the regular tree languages---and
 therefore the context-free string languages by the fact cited
above---are exactly those languages that are definable within the
weak monadic second-order logic with multiple successors. These
definability results can easily be extended to context-free tree
languages by lifting their alphabets and eliminating in this
process all their nonterminal symbols of arity greater than zero.
Given the close relationship between tag's and monadic context-free
tree grammars, established in this note, it is possible to apply
the definitional resources of the weak monadic second-order logic
to tag's and to give grammar independent characterizations of this
formalism.

As another result of the equivalence between tag's and context-free
tree grammars, parsing techniques and special concepts of finite
automata that were developed for cftg's, become immmediately
applicable to tag's. It remains to be investigated whether these
methods from a different type of grammatical formalism can advance
the theory of tree adjoining grammars.


\begin{thebibliography}{}

\bibitem[Engelfriet and Schmidt, 1977]{ES77:IO-OI1}
Engelfriet, J. and Schmidt, E.~M. (1977).
\newblock {IO} and {OI}, part {I}.
\newblock {\em J. Comput. System Sci.}, 15:328--353.

\bibitem[Fischer, 1968]{fischer68}
Fischer, M.~J. (1968).
\newblock Grammars with macro-like productions.
\newblock In {\em Proceedings of the 9th Annual Symposium on Switching and
  Automata Theory}, pages 131--142. IEEE.

\bibitem[Maibaum, 1974]{maibaum74}
Maibaum, T. S.~E. (1974).
\newblock A generalized approach to formal languages.
\newblock {\em J. Comput. System Sci.}, 8:409--439.

\bibitem[Vijay-Shanker and Weir, 1992]{vijay-shankerWeir92}
Vijay-Shanker, K. and Weir, D. (1992).
\newblock The equivalence of four extensions of context-free grammars.
\newblock Cognitive Science Research Paper 236, Univ. of Essex.

\end{thebibliography}

\end{document}